\documentclass[sigconf]{acmart}
\copyrightyear{2021} 
\acmYear{2021} 
\setcopyright{acmlicensed}
\acmConference[MM '21]{Proceedings of the 29th ACM International Conference on Multimedia}{October 20--24, 2021}{Virtual Event, China}
\acmBooktitle{Proceedings of the 29th ACM International Conference on Multimedia (MM '21), October 20--24, 2021, Virtual Event, China}
\acmPrice{15.00}
\acmDOI{10.1145/3474085.3478329}
\acmISBN{978-1-4503-8651-7/21/10}

\title{PyTorchVideo: \\A Deep Learning Library for Video Understanding}

\usepackage[utf8]{inputenc}
\usepackage[english]{babel}
\usepackage{algorithm}
\usepackage{listings}

\def\BibTeX{{\rm B\kern-.05em{\sc i\kern-.025em b}\kern-.08emT\kern-.1667em\lower.7ex\hbox{E}\kern-.125emX}}

\acmSubmissionID{3254}

\newcommand{\customfootnotetext}[2]{{
  \renewcommand{\thefootnote}{#1}
  \footnotetext[0]{#2}}}
\begin{document}

\author{Haoqi Fan$^*$, Tullie Murrell$^*$, Heng Wang$^\dagger$, Kalyan Vasudev Alwala$^\dagger$,\\ Yanghao Li$^\dagger$, Yilei Li$^\dagger$, \mbox{Bo Xiong$^\dagger$}, Nikhila Ravi, Meng Li, Haichuan Yang,\\  Jitendra Malik, Ross Girshick, Matt Feiszli, Aaron Adcock$^\ddagger$, Wan-Yen Lo$^\ddagger$, Christoph Feichtenhofer$^\ddagger$} 

\affiliation{%
  \institution{Facebook AI}
}

\renewcommand{\shortauthors}{Fan, et al.}

\begin{abstract}

We introduce PyTorchVideo, an open-source deep-learning library that provides a rich set of modular, efficient, and reproducible components for a variety of video understanding tasks, including classification, detection, self-supervised learning, and low-level processing. The library covers a full stack of video understanding tools including multimodal data loading, transformations, and models that reproduce state-of-the-art performance. PyTorchVideo further supports hardware acceleration that enables real-time inference on mobile devices. The library is based on PyTorch and can be used by any training framework; for example, PyTorchLightning, PySlowFast, or Classy Vision. PyTorchVideo is available at \url{https://pytorchvideo.org/}

\customfootnotetext{* $^\dagger$ $\ddagger$}{Indicating equal technical contribution.}

\end{abstract}



\begin{teaserfigure}
  \includegraphics[width=\textwidth]{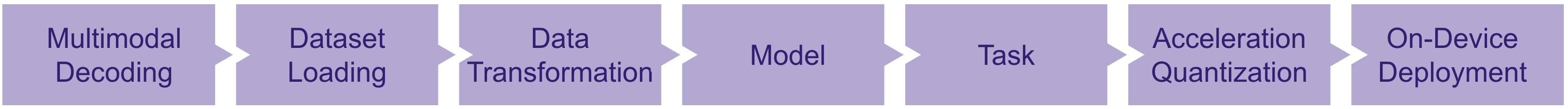}
  \vspace{-18pt}
  \caption{A list of full stack video understanding components provided by PyTorchVideo.}
  \label{fig:teaser}
\end{teaserfigure}

\maketitle

\section{Introduction}

 Recording, storing, and viewing videos has become an ordinary part of our lives; in 2022, video traffic will amount to \textit{82 percent of all Internet traffic}~\cite{cisco_christoph}. With the increasing amount of video material readily available (e.g. on the web), it is now more important than ever to develop ML frameworks for video understanding. 

With the rise of deep learning, significant progress has been made in video understanding research, with novel neural network architectures, better training recipes, advanced data augmentation, and model acceleration techniques. However, the sheer amount of data that video brings often makes these tasks computationally demanding; therefore efficient solutions are non-trivial to implement.

To date, there exist several popular video understanding frameworks, which provide implementations of advanced state-of-the-art video models, including PySlowFast~\cite{fan2020pyslowfast}, MMAction~\cite{mmaction2019}, MMAction2 \cite{2020mmaction2}, and Gluon-CV~\cite{gluoncvnlp2020}. However, unlike a modularized \emph{library} that can be imported into different projects, all of these frameworks are designed around training workflows, which limit their adoption beyond applications tailored to one specific codebase. 

More specifically, we see the following limitations in prior efforts. 
First, reproducibility -- an important requirement for deep learning software -- varies across frameworks; e.g. identical models are reproduced with varying accuracy on different frameworks~\cite{fan2020pyslowfast,mmaction2019,2020mmaction2,gluoncvnlp2020}. Second, regarding input modalities, the frameworks are mainly focused on visual-only data streams.
Third, supported tasks only encompass human action classification and detection. Fourth, none of the existing codebases support on-device acceleration for real-time inference on mobile hardware.

We believe that a modular, \textit{component-focused} video understanding library that addresses the aforementioned limitations will strongly support the video research community. Our intention is to develop a library that aims to provide fast and easily extensible components to benefit researchers and practitioners in academia and industry. 

We present PyTorchVideo -- an efficient, modular and reproducible deep learning library for video understanding which supports the following (see Fig.~\ref{fig:teaser} for an overview): 
\vspace{-0.5em}
\begin{itemize}
\item a \textbf{modular} design with extendable interface for video modeling using Python
\item a \textbf{full stack} of video understanding machine learning components from established datasets to state-of-the-art models
\item \textbf{real-time} video classification through hardware accelerated \textbf{on-device} support
\item multiple \textbf{tasks}, including human action classification and detection, self-supervised learning, and low-level vision tasks
\item \textbf{reproducible} models and datasets, benchmarked in a comprehensive model zoo 
\item  multiple input \textbf{modalities}, including visual, audio, optical-flow and IMU data
\end{itemize}
\vspace{-3pt}
PyTorchVideo is distributed with a Apache 2.0 License, and is available on GitHub at \url{https://github.com/facebookresearch/pytorchvideo}.

\vspace{-3pt}

\section{Library Design}
Our library follows four design principles, outlined next (\S\ref{sec:modularity}-\ref{sec:reproducibility}). \vspace{-7pt}

\subsection{Modularity} \label{sec:modularity}

PyTorchVideo is built to be component centric: it provides independent components that are plug-and-play and ready to mix-and-match for any research or production use case. We achieve this by designing models, datasets and data transformations (transforms) independently, only enforcing consistency through general argument naming guidelines. For example, in the \verb|pytorchvideo.data| module all datasets provide a \verb|data_path| argument, or, for the \verb|pytorchvideo.models| module, any reference to input dimensions uses the name \verb|dim_in|. This form of duck-typing provides flexibility and straightforward extensibility for new use cases.

\vspace{-7pt}
\subsection{Compatibility}

PyTorchVideo is designed to be compatible with other frameworks and domain specific libraries. In contrast to existing video frameworks~\cite{fan2020pyslowfast,mmaction2019,2020mmaction2,gluoncvnlp2020}, PyTorchVideo does not rely on a configuration system. To maximize the compatibility with Python based frameworks that can have arbitrary config-systems, PyTorchVideo uses keyword arguments in Python as a ``naive configuration'' system. 

PyTorchVideo is designed to be interoperable with other standard domain specific libraries by setting canonical modality based tensor types. For videos, we expect a tensor of shape $[..., C, T, H, W]$, where $T\times H\times W$ are spatiotemporal dimensions, and $C$ is the number of color channels, allowing any TorchVision model or transform to be used together with PyTorchVideo. For raw audio waveforms, we expect a tensor of shape $[..., T]$, where $T$ is the temporal dimension, and for spectrograms, we expect a tensor of shape $[..., T, F]$, where $T$ is time and $F$ is frequency, aligning with TorchAudio.

\subsection{Customizability}

\begin{algorithm}[t]
\caption{Code for a SlowFast network with customized norm and activation layer classes, and a custom head function.}
\label{code:customize-model}

\definecolor{codeblue}{rgb}{0.25,0.5,0.5}
\lstset{
  backgroundcolor=\color{white},
  basicstyle=\fontsize{7.2pt}{7.2pt}\ttfamily\selectfont,
  columns=fullflexible,
  breaklines=true,
  captionpos=b,
  commentstyle=\fontsize{7.2pt}{7.2pt}\color{codeblue},
  keywordstyle=\fontsize{7.2pt}{7.2pt},
}
\begin{lstlisting}[language=python]
    # Create a customized SlowFast Network.
    customized_slowfast = create_slowfast(
        activation=CustomizedActivation,
        norm=CustomizedNorm, 
        head=create_customized_head, 
    ) 
\end{lstlisting}
\end{algorithm}

One of PyTorchVideo's primary use cases is supporting the latest research methods; we want researchers to easily contribute their work without requiring refactoring and architecture modifications. To achieve this, we designed the library to reduce overhead for adding new components or sub-modules. Notably in the \verb|pytorchvideo.model| module, we use a dependency injection inspired API. We have a composable interface, which contains injectable skeleton classes and a factory function interface that builds reproducible implementations using composable classes. We anticipate this injectable class design to being useful for researchers that want to easily plug in new sub-components (e.g. a new type of convolution) into the structure of larger models such as a ResNet~\cite{He2016} or SlowFast~\cite{Feichtenhofer2019}. The factory functions are more suitable for reproducible benchmarking of complete models, or usage in production. An example for a customized SlowFast network is in Algorithm~\ref{code:customize-model}.

\subsection{Reproducibility} \label{sec:reproducibility}

PyTorchVideo maintains reproducible implementations of all models and datasets. Each component is benchmarked against the reported performance in the respective, original publication. We report performance and release model files online\footnote{Numbers and model weights can be found in \url{https://github.com/facebookresearch/pytorchvideo/blob/master/docs/source/model_zoo.md}} as well as on PyTorch Hub\footnote{\url{https://pytorch.org/hub/}}. We rely on test coverage and recurrent benchmark jobs to verify and monitor performance and to detect potential regressions introduced by codebase updates.

\section{Library Components}

PyTorchVideo allows training of state-of-the-art models on multi-modal input data, and deployment of an accelerated real-time model on mobile devices. Example components are shown in Algorithm~\ref{alg:code}.

\begin{algorithm}[t]
\caption{Code snippet to train, run inference, and deploy a state-of-the-art video model with PyTorchVideo.}
\label{alg:code}

\definecolor{codeblue}{rgb}{0.25,0.5,0.5}
\lstset{
  backgroundcolor=\color{white},
  basicstyle=\fontsize{7.2pt}{7.2pt}\ttfamily\selectfont,
  columns=fullflexible,
  breaklines=true,
  captionpos=b,
  commentstyle=\fontsize{7.2pt}{7.2pt}\color{codeblue},
  keywordstyle=\fontsize{7.2pt}{7.2pt},
}
\begin{lstlisting}[language=python]
    from pytorchvideo import data, models, accelerator
    # Create visual and acoustic models.
    visual_model = models.slowfast.create_slowfast(
        model_num_class=400,
    ) 
    acoustic_model = models.resnet.create_acoustic_resnet(
        model_num_class=400,
    ) 
    # Create Kinetics dataloader. 
    kinetics_loader = torch.utils.data.DataLoader(
        data.Kinetics(
            data_path=DATA_PATH,
            clip_sampler=data.make_clip_sampler(
                "uniform", 
                CLIP_DURATION,
            ),
        )
        batch_size=BATCH_SIZE,
    ) 
    # Deploy model.
    visual_net_inst_deploy = accelerator.deployment.\ 
        convert_to_deployable_form(net_inst, input_tensor)

\end{lstlisting}
\end{algorithm}

\subsection{Data}

Video contains rich information streams from various sources, and, in comparison to image understanding, video is more computationally demanding. \mbox{PyTorchVideo} provides a modular, and efficient data loader to decode visual, motion (optical-flow), acoustic, and Inertial Measurement Unit (IMU) information from raw video.

PyTorchVideo supports a growing list of data loaders for various popular video datasets for different tasks: video classification task for UCF-101~\cite{Soomro2012}, HMDB-51~\cite{Kuehne2011}, Kinetics~\cite{Kay2017}, Charades~\cite{Sigurdsson2016}, and Something-Something~\cite{ssv2}, egocentric tasks for Epic Kitchen~\cite{Damen2018EPICKITCHENS} and DomSev~\cite{Silva2018}, as well as video detection in AVA~\cite{Gu2018}.

All data loaders support several file formats and are data storage agnostic. For encoded video datasets (e.g. videos stored in mp4 files), we provide PyAV, TorchVision, and Decord decoders. For long videos -- when decoding is an overhead -- PyTorchVideo provides support for pre-decoded video datasets in the form image files.

\subsection{Transforms}

Transforms - as the key components to improve the model generalization - are designed to be flexible and easy to use in PyTorchVideo. PyTorchVideo provides factory transforms that include common recipes for training state-of-the-art video models~\cite{Feichtenhofer2019,feichtenhofer2020x3d,mvit}. Recent data augmentations are also provided by the library (e.g. MixUp, CutMix, RandAugment~\cite{cubuk2020randaugment}, and AugMix~\cite{hendrycks2020augmix}). Finally, users have the option to create custom transforms by composing individual ones.

\subsection{Models}

PyTorchVideo contains highly reproducible implementations of popular models and backbones for video classification, acoustic event detection, human action localization (detection) in video, as well as self-supervised learning algorithms.

The current set of models includes  standard single stream video backbones such as C2D~\cite{Wang2018}, I3D~\cite{Wang2018}, Slow-only~\cite{Feichtenhofer2019} for RGB frames and acoustic ResNet~\cite{fanyi2020} for audio signal, as well as efficient video networks such as SlowFast~\cite{Feichtenhofer2019}, CSN~\cite{tran2019}, R2+1D~\cite{tran2018}, and X3D~\cite{feichtenhofer2020x3d} that provide state-of-the-art performance. PyTorchVideo also provides multipathway architectures such as Audiovisual SlowFast networks~\cite{fanyi2020} which enable state-of-the-art performance by disentangling spatial, temporal, and acoustic signals across different pathways. 

  It further supports methods for low-level vision tasks for researchers to build on the latest trends in video representation learning.

PyTorchVideo models can be used in combination with different downstream tasks: {supervised} classification and detection of human actions in video~\cite{Feichtenhofer2019}, as well as self-supervised (i.e. unsupervised) video representation learning with Momentum Contrast~\cite{moco}, \mbox{SimCLR}~\cite{simclr}, and Bootstrap your own latent~\cite{byol}.

\subsection{Accelerator}

\begin{figure}[t]
  \includegraphics[width=0.47\textwidth]{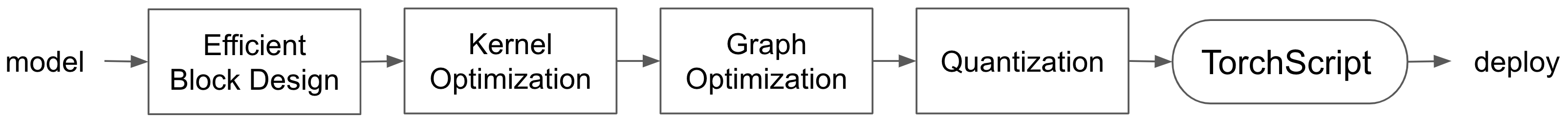}
  \vspace{-1em}

  \caption{Acceleration and deployment pipeline.}
  \Description{}
  \label{fig:acceleration}
\vspace{-.5em}
\end{figure}

PyTorchVideo provides a complete environment (Accelerator) for hardware-aware design and deployment of models for fast inference on device, including efficient blocks and kernel optimization. The deployment flow is illustrated in Figure \ref{fig:acceleration}.

Specifically, we perform kernel-level latency optimization for common kernels in video understanding models (e.g. conv3d). This optimization brings two-fold benefits: (1) latency of these kernels for floating-point inference is significantly reduced; (2) quantized operation (int8) of these kernels is enabled, which is not supported for mobile devices by vanilla PyTorch. Our Accelerator provides a set of efficient blocks that build upon these optimized kernels, with their low latency validated by on-device profiling. In addition, our Accelerator provides kernel optimization in deployment flow, which can automatically replace modules in the original model with efficient blocks that perform equivalent operations. Overall, the PyTorchVideo Accelerator provides a complete environment for hardware-aware model design and deployment for \textit{fast} inference.

\section{Benchmarks}
PyTorchVideo supports popular video understanding tasks, such as video classification~\cite{Kay2017,ssv2,Feichtenhofer2019}, action detection~\cite{Gu2018}, video self-supervised learning~\cite{feichtenhofer2021large} and efficient video understanding on mobile hardware. We provide comprehensive benchmarks for different tasks and a large set of model weights for state-of-the-art methods. This section provides a snapshot of the benchmarks. A comprehensive listing can be found in the model zoo \footnote{A full set of supported models on different datasets can be found from \url{https://pytorchvideo.readthedocs.io/en/latest/model_zoo.html}}. The models in PyTorchVideos' model zoo reproduce performance of the original publications and can seed future research that builds upon them, bypassing the need for re-implementation and costly re-training. 
\subsection{Video classification}

\begin{table}[h!]
    \small
	\centering
	\setlength{\tabcolsep}{3.5pt}
	
	\begin{tabular}{lcccc}
			\toprule

        model & Top-1  & Top-5  & FLOPs (G)   & Param (M) \\

		\hline
		I3D R50, 8$\times$8~\cite{Wang2018} &  73.3 & 90.7 & 37.5 x 3 x 10 & 28.0 \\

 		X3D-L, 16$\times$5~\cite{feichtenhofer2020x3d} &  77.4 &	93.3&	26.6 x 3 x 10	&6.2 \\	
 		SlowFast R101, 16$\times$8~\cite{Feichtenhofer2019} &  78.7&	93.6&	215.6 x 3 x 10	&53.8 \\
 		MViT-B, 16$\times$4~\cite{mvit} & 78.8 & 93.9	&	70.5 × 3 × 10 & 36.6 \\
	    \bottomrule
	\end{tabular}
	\caption{Video classification results on Kinetics-400~\cite{Kay2017}. 
	}\label{tab:sota:k400}
	\vspace{-2.5em}
\end{table}

PyTorchVideo implements classification for various datasets, including UCF-101~\cite{Soomro2012}, HMDB-51~\cite{Kuehne2011}, Kinetics~\cite{Kay2017}, Charades~\cite{Sigurdsson2016}, Something-Something~\cite{ssv2} and Epic-Kitchens~\cite{Damen2018EPICKITCHENS}. Table~\ref{tab:sota:k400} shows a benchmark snapshot on Kinetics-400 for three popular state-of-the-art methods, which are measured in Top-1 and Top-5 accuracies. Further classification models with pre-trained weights, that can be used for a variety of downstream tasks, are available online$^\text{3}$.

\subsection{Video action detection}

The video action detection task aims to perform spatiotemporal localization of human actions in videos. Table~\ref{tab:sota:ava} shows detection performance in mean Average Precision (mAP) on the AVA dataset~\cite{Gu2018} using Slow and SlowFast networks~\cite{Feichtenhofer2019}.

\begin{table}[t!]
    \small
	\centering
	\setlength{\tabcolsep}{4.5pt}
	
	\begin{tabular}{lccc}
			\toprule

        model & mAP  & Param (M) \\

		\hline
 		Slow R50, 4$\times$16~\cite{Feichtenhofer2019} & 19.50 & 31.78 \\
 		SlowFast R50, 8$\times$8~\cite{Feichtenhofer2019} &  24.67 &	33.82 \\
	    \bottomrule
	\end{tabular}
	\caption{\textbf{Video action detection results on AVA dataset ~\cite{Gu2018}}. 
	}\label{tab:sota:ava}
	\vspace{-2.5em}
\end{table}

\subsection{Video self-supervised learning}

We provide reference implementations of popular self-supervised learning methods for video~\cite{feichtenhofer2021large}, which can be used to perform unsupervised spatiotemporal representation learning on large-scale video data. Table~\ref{tab:sota:ssl} summarizes the results on 5 downstream datasets.

\begin{table}[t!]
    \small
	\centering
	\setlength{\tabcolsep}{3.5pt}

	\begin{tabular}{lccccc}
	    \toprule

		SSL method & Kinetics & UCF101 & AVA & Charades   & SSv2
		\\
		
		\hline
		SimCLR~\cite{simclr} &  62.0 & 87.9 & 17.6 & 11.4 & 52.0 \\		
		BYOL~\cite{byol} &  68.3 & 93.8 & 23.4 & 21.0 & 55.8 \\		
		MoCo~\cite{moco} &  67.3 & 92.8 & 20.3 & 33.5 & 54.4 \\	
		\bottomrule
	\end{tabular}
	\caption{Performance of 4 video self-supervised learning (SSL) methods on 5 downstream datasets~\cite{feichtenhofer2021large}.
	}\label{tab:sota:ssl}

	\vspace{-2.5em}
\end{table}

\subsection{Efficient mobile video understanding}
The Accelerator environment for efficient video understanding on mobile devices is benchmarked in Table~\ref{tab:sota:mobile}. We show several efficient X3D models~\cite{feichtenhofer2020x3d} on a Samsung Galaxy S9 mobile phone. With the efficient optimization strategies in PyTorchVideo, X3D achieves {4.6 - 5.6$\times$} inference speed up, compared to the default PyTorch implementation. With quantization (int8), it can be further accelerated by {1.4$\times$}. The resulting PyTorchVideo-accelerated X3D model runs around \textbf{6}$\times$ \textit{faster than real time}, requiring roughly 165ms to process one second of video, \textit{directly on the mobile phone}. The source code for an on-device demo in iOS\footnote{\url{https://github.com/pytorch/ios-demo-app/tree/master/TorchVideo}} and Android\footnote{\url{https://github.com/pytorch/android-demo-app/tree/master/TorchVideo}} is available.

\begin{table}[t!]
    \small
	\centering
	\setlength{\tabcolsep}{3.5pt}
	\label{tab:sota:in1k}
	\begin{tabular}{lccc}
	    \toprule

		model 

		& \multicolumn{1}{p{2cm}}{\centering Vanilla PyTorch \\ Latency (ms) } 
		& \multicolumn{1}{p{2cm}}{\centering PyTorchVideo \\ Latency (ms) }  
		& Speed Up   
		\\
	
		\hline
		{X3D-XS (fp32)~\cite{feichtenhofer2020x3d}} &    1067 & 233 & 4.6$\times$  \\
		{X3D-S (fp32)~\cite{feichtenhofer2020x3d}} &  4249 & 764 & 5.6$\times$  \\
		{X3D-XS (int8)~\cite{feichtenhofer2020x3d}} &    (not supported) & 165 & -  \\
		\bottomrule
	\end{tabular}
	\caption{\textbf{Speedup of PyTorchVideo models on mobile CPU.} \label{tab:sota:mobile}
	}
	\vspace{-2.5em}
\end{table}

\section{Conclusion}

We introduce PyTorchVideo, an efficient, flexible, and modular deep learning library for video understanding that scales to a variety of research and production applications. Our library welcomes contributions from the research and open-source community, and will be continuously updated to support further innovation. In future work, we plan to continue enhancing the library with optimized components and reproducible state-of-the-art models. 

\begin{acks}
We thank the PyTorchVideo contributors:  Aaron Adcock, Amy Bearman, Bernard Nguyen, Bo Xiong, Chengyuan Yan, Christoph Feichtenhofer, Dave Schnizlein, Haoqi Fan, Heng Wang, Jackson Hamburger, Jitendra Malik, Kalyan Vasudev Alwala, Matt Feiszli, Nikhila Ravi, Ross Girshick, Tullie Murrell, Wan-Yen Lo, Weiyao Wang, Yanghao Li, Yilei Li, Zhengxing Chen, Zhicheng Yan.
\end{acks}

\bibliographystyle{ACM-Reference-Format}
\bibliography{arxiv}

\end{document}